\documentclass[conference]{IEEEtran}
\IEEEoverridecommandlockouts
\usepackage{cite}
\usepackage{amsmath,amssymb,amsfonts}
\usepackage{algorithmic}
\usepackage{graphicx}
\usepackage{textcomp}

\usepackage{natbib}
\setcitestyle{numbers,square}

\usepackage{CJKutf8}

\usepackage{multirow}
\usepackage[table,xcdraw]{xcolor}
\usepackage{graphicx}
\usepackage{makecell}
\usepackage{booktabs}
\usepackage[justification=centering]{caption}

\begin{document}

\title{Application and Optimization of Large Models Based on Prompt Tuning for Fact-Check-Worthiness Estimation\\
}

\author{\IEEEauthorblockN{
		Yinglong Yu \IEEEauthorrefmark{1}\textsuperscript{1}, 
		Hao Shen \IEEEauthorrefmark{1}\textsuperscript{2}
		Zhengyi Lyu\IEEEauthorrefmark{1}\textsuperscript{3}
            and Qi He\IEEEauthorrefmark{1}\textsuperscript{4}}
            
        \IEEEauthorblockA{\IEEEauthorrefmark{1}Communication University of China, Beijing, China}
		\textsuperscript{1}yuyingling@cuc.edu.cn
            \textsuperscript{2}shenhao@cuc.edu.cn
            \textsuperscript{3}lyuzhengyi@cuc.edu.cn
            \textsuperscript{4}heqi654321@126.com
            \thanks{Hao Shen is corresponding author.}
        }


\maketitle

\begin{abstract}
In response to the growing problem of misinformation in the context of globalization and informatization, this paper proposes a classification method for fact-check-worthiness estimation based on prompt tuning. We construct a model for fact-check-worthiness estimation at the methodological level using prompt tuning. By applying designed prompt templates to large language models, we establish in-context learning and leverage prompt tuning technology to improve the accuracy of determining whether claims have fact-check-worthiness, particularly when dealing with limited or unlabeled data. Through extensive experiments on public datasets, we demonstrate that the proposed method surpasses or matches multiple baseline methods in the classification task of fact-check-worthiness estimation assessment, including classical pre-trained models such as BERT, as well as recent popular large models like GPT-3.5 and GPT-4. Experiments show that the prompt tuning-based method proposed in this study exhibits certain advantages in evaluation metrics such as F1 score and accuracy, thereby effectively validating its effectiveness and advancement in the task of fact-check-worthiness estimation.
\end{abstract}


\section{Introduction}
In today’s interconnected world characterized by globalization and informatization, the complexity of multilingual environments and the challenges posed by misinformation have become increasingly severe. With the deepening of international exchanges and the expanding influence of social media, rumors and false information spread rapidly across cyberspace, exacerbating the uncertainty in the global public discourse.

Fact-checking tasks generally refer to verifying and confirming specific statements or claims. This process may involve collecting, analyzing, comparing, and verifying data to ensure that the provided information is accurate, reliable, and valid. In existing research on fact-checking, the information to be verified is referred to as a claim \citep{zhou2020survey}. The typical workflow of a fact-checking task involves assessing the claim's veracity. In contrast, the task of fact-check-worthiness estimation, which is the precursor to fact-checking, primarily aims to determine whether a piece of textual information is worthy of fact-checking.

With the development of large language models, uncovering the latent knowledge within these models has become one of the current hot topics. Prompt tuning, through the design of ingenious templates that guide the underlying model to adapt to downstream tasks, is considered capable of fully mobilizing the model's internal knowledge. It has been proven to perform well in many natural language processing tasks \citep{liu2023pre}.

This paper attempts to apply prompt tuning to fact-check-worthiness estimation, proposing an evaluation classification method based on prompt tuning. This method combines the design of prompt templates with in-context learning and prompt tuning to collaboratively enhance the potential of large language models in understanding fact-check-worthiness estimation tasks. The effectiveness of the proposed method is evaluated on public datasets, and experimental results indicate that its performance on fact-check-worthiness estimation tasks surpasses classical classification methods and is comparable to that of the GPT-4 model.

\section{Related Work}
\subsection{Fact-checking}

Fact-checking is inseparably linked with public opinion analysis; some traditional public opinion analysis companies hire specialized fact-checkers to analyze and verify the accuracy of factual claims in news articles or social media information. Initially, fact-checking focused primarily on political statements and campaign advertisements. With the development of internet social media and the proliferation of user-generated content, the importance of fact-checking has become increasingly significant.

Recent advancements in deep learning technologies have increased interest among researchers in applying these techniques to automated fact-checking systems. Stammbach and Ash \cite{stammbach2020fever} utilized LSTM combined with external text to classify claim texts. Lewis et al. and Maillard et al. \cite{lewis2020retrieval,maillard2021multi} improved fact-checking evidence retrieval through neural networks and dot product indexing. Text generation technologies also play a role in this field; for example, the QACG project uses language models to generate verification models for claims \cite{pan2021zero}, while Zuo et al. \cite{zuo2018hybrid} leveraged GPT-3 to create summaries of verification evidence. Additionally, prompt engineering, such as PEINet, which combines prompts with recursive graph attention networks, can effectively verify multilingual claims \citep{li2022peinet}.

\subsection{Fact-check-worthiness Estimation}

The task of fact-check-worthiness estimation is a subtask of fact-checking to determine whether a given text is worth fact-checking. This task serves as the starting point for fact-checking; subsequent fact-checking steps can be skipped if a text isn’t worth fact-checking. Typically, this task relies on professional fact-checkers for verification or involves assessing the text through auxiliary questions.

\subsection{Prompt Tuning}

Since the introduction of models such as BERT \citep{devlin2018bert}, GPT \citep{radford2018improving}, and ELMo \citep{peters2018deepcontextualizedwordrepresentations}, the working paradigm for natural language processing tasks has entered the era of `pre-training + fine-tuning.' The concept of prompts was introduced to address issues with the `pre-training + fine-tuning' approach, such as introducing new parameters at the model's head during the fine-tuning stage and overfitting in few-shot or zero-shot scenarios. The concept of prompt tuning originated with GPT-3 \citep{brown2020language}, which posited that providing large-scale models with specific templates could significantly enhance their capabilities in terms of understanding and reasoning.

In recent years, an increasing number of researchers have applied prompt tuning tasks to a wide range of natural language processing tasks. Prompt templates can be relatively quickly constructed for these tasks, and the key to solving these problems lies in defining a suitable template. The definition of templates varies widely. For instance, Yin et al. \cite{yin2019benchmarking}, to achieve text topic classification, defined the template as `The topic of this document is [mask],' transforming the approach to solving text classification problems into a text generation task. Kojima et al. \cite{kojima2022large}, to improve logical reasoning issues such as arithmetic and symbolic reasoning in few-shot learning scenarios with large language models, defined the template as `Let’s think step by step,' guiding the model to think through problems step-by-step. In the field of text classification, most researchers use cloze-style prompts to create templates and have extensively explored the design of prompt templates.

\subsection{Summary}

Upon analyzing and summarizing existing work, it can be concluded that: 

\begin{itemize}

\item  While current fact-checking technologies and fact-check-worthiness estimation techniques have made significant progress, they still possess limitations. Some models applied to fact-checking tasks continue the traditional training paradigm of `pre-training + fine-tuning.' Although this supervised learning framework is practical to a certain extent, the conclusions generated often lack intrinsic logical analysis and transparency. 

\item  Fact-check-worthiness estimation requires specialized background knowledge and substantial common-sense experience. Customized prompt templates for downstream tasks have shown considerable potential in practical applications, effectively activating and utilizing the rich human knowledge embedded within large language models. This enhances and supports the judgment process of fact-check-worthiness, thereby improving assessments' overall accuracy and reliability.

\end{itemize}

\section{Method}

\subsection{Task Definition}

This model aims to determine whether a claim is worthy of fact-checking. This paper describes the mathematical formulation of the fact-check-worthiness estimation problem. For a dataset $T=\{X,Y\}$, where $X=\{x_1,x_2,\ldots,x_n\}$ and $Y=\{y_1,y_2,\ldots,y_m\}$,this indicates that the dataset contains $n$ texts and $m$ judgment labels.The model's output is the label $y_j \in Y$ for each text $x_i$.The objective of this study is to train a model that improves the accuracy of the prediction function, ensuring that the predicted labels $y_j$ match the true labels $y_i$, The specific equation is as follows:
\begin{equation}\label{eq1}
y_{j}=\operatorname{argmax}_{y \in Y} P(y \mid x)
\end{equation}

\subsection{Prompt Template Design}

Prompt tuning focuses on constructing effective templates, thereby converting the fact-check-worthiness estimation classification task into a text generation task. This paper designs two templates for the final response, tailored to the usage scenario. The specific templates are provided in Table \ref{tab:1}.

\begin{table*}[htbp]
\caption{Two types of prompt templates.}
\begin{center}
\renewcommand{\arraystretch}{1.5}
\begin{tabular}{|c|c|}
\hline
\textbf{Name} & \textbf{Content} \\ 
\hline
\multirow{5}{*}{Long template} & \multirow{5}{*}{\makecell[c]{The aim of this task is to determine whether a claim in a tweet and/or transcriptions is worth \\ fact-checking. Typical approaches to make that decision require to either resort to the \\ judgments of professional fact-checkers or to human annotators to answer several auxiliary \\ questions such as ``does it contain a verifiable factual claim?", and ``is it harmful?", before \\ deciding on the final check-worthiness label.You just need to answer "Yes" or "No".}} \\ 
& \\
& \\
& \\
& \\
\hline
\multirow{2}{*}{Short template} & \multirow{2}{*}{\makecell[c]{Is the claim in this tweet worth conducting factual \\  verification? You just need to answer ``Yes" or "No".}} \\ 
& \\
\hline
\end{tabular}
\end{center}
\label{tab:1}
\end{table*}

In the designed prompt template structure, the `\textbf{Task: + Template + Text: + Claim + \textbackslash n +[mask]}' format is adopted, where \textbf{[mask]} serves as a placeholder symbol representing the core content of the prediction to be generated by the model. To achieve dual optimization goals, this paper employs two prompt templates of different lengths for the following reasons: 
\begin{itemize}

\item By using templates of varying lengths, the aim is to circumvent the maximum input token limit of the underlying language model, thereby avoiding the omission of critical text information or important prompt phrases due to exceeding the token limit. 

\item This strategy also aims to reduce the impact of redundant information on the model's decision-making process, preventing smaller-parameter models from being inundated with irrelevant information when using longer templates, which could otherwise impair their accuracy.

\end{itemize}

The longer template provides further clarification on identifying the source of the text, describes the workflow of this task, and offers two criteria for judgment. Compared to the shorter template, we believe these descriptions can better elicit the potential of large language models in fact-checking. Additionally, both template structures include a clear constraint at the end: `\textbf{You just need to answer Yes or No.}' This setting aims to restrict the model's output to ensure it does not generate extraneous responses that would be difficult to precisely map to predefined categories in subsequent stages of answer processing.

\subsection{Verbalizer Design}

The goal of the verbalizer is to map the predicted words generated by the model to the required target categories. Although this paper includes constraints in prompt engineering, the results generated by large language models remain unstable, necessitating the design of an answer engineering process to convert the generated text into the desired outcomes for fact-check-worthiness estimation. To reduce computational resource consumption and considering the characteristics of this task, this paper designs a simple and straightforward answer engineering transformation formula:

\begin{equation}\label{eq2}
\mathrm A_{\mathrm f}=\left\{\begin{matrix}\text{ Yes, If 'es' in $\mathrm A_{\mathrm i}$}\\\text{ No, else}\end{matrix}\right.
\end{equation}

Here, $\mathrm A_{\mathrm f}$ represents the result after answer engineering transformation, and $\mathrm A_{\mathrm i}$represents the raw output generated by the model. Since the answers typically only contain 'Yes' or 'No,' if the string 'es' appears consecutively in the generated result, it is classified as 'Yes'; otherwise, it is classified as 'No.'

\subsection{In-Context Learning Method Design}

In-context learning involves selecting a few labeled samples from the training set and inserting these samples into the prompt template design to guide the model in generating the appropriate results. This approach essentially instructs the model on how to solve the task. These training samples and their corresponding labels are referred to as Demonstrations. This method is subsequently referred to as \textbf{n-shot} in this paper, where \textbf{$n$} represents the number of labeled samples chosen for assistance.

\begin{figure*}[]
	\centering
	\includegraphics[width=.9\textwidth]{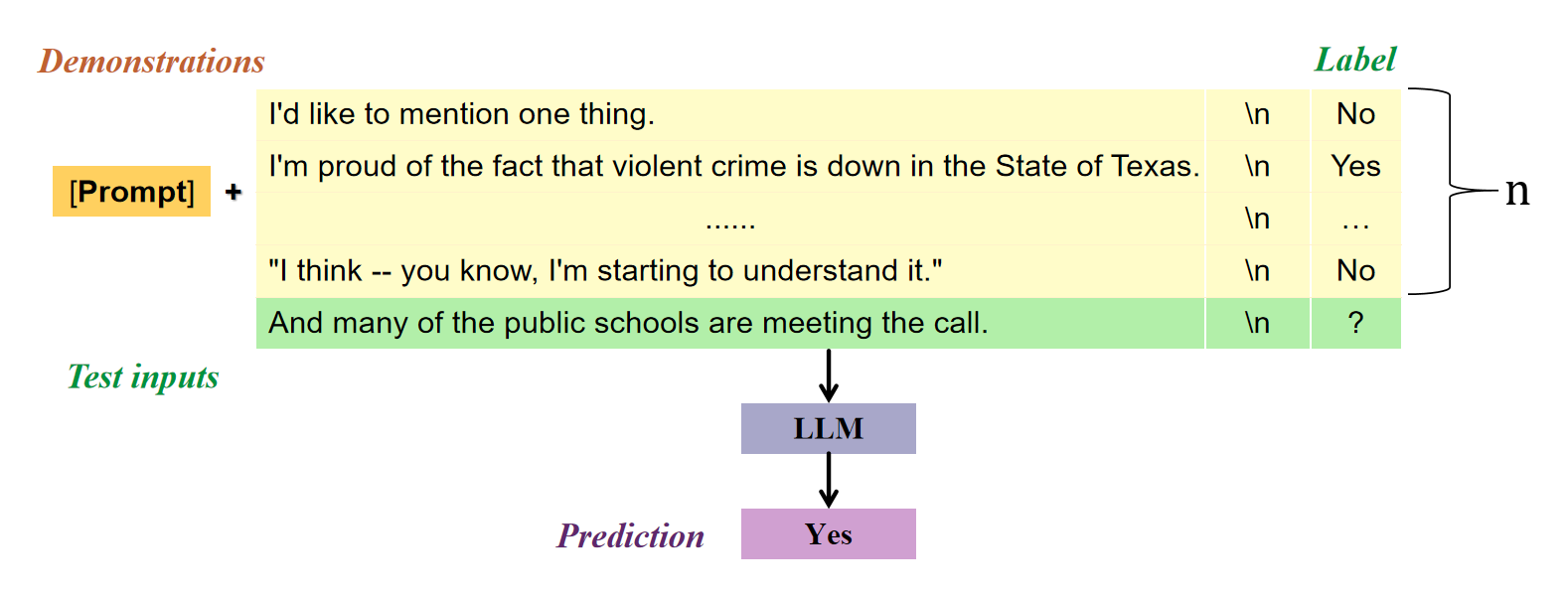}
	\caption{Workflow of in-context learning method.}
	\label{fig:1}
\end{figure*}

Figure \ref{fig:1} illustrates the workflow of In-Context Learning. Each time, $n$ training samples are randomly selected as Demonstrations, along with their corresponding labels. The 'Prompt' refers to the prompt template, and \textbackslash n is used to distinguish between the sentences that need to be judged and their labels. The green parts represent the test samples. Finally, this text is input into the large language mode, which outputs the judgment result for fact-checking.

\subsection{Prompt Engineering Design}

When designing prompt engineering, it is necessary to skillfully combine the prompt templates developed above, verbalizer, and in-context learning methods.

\begin{figure*}[]
	\centering
	\includegraphics[width=.9\textwidth]{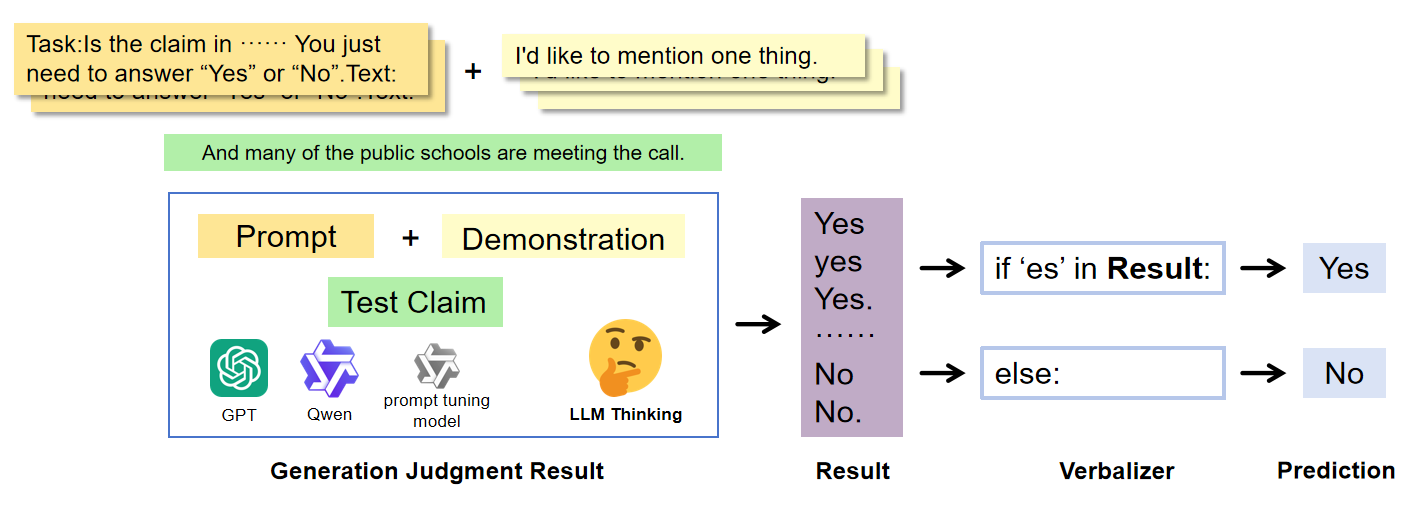}
	\caption{Workflow of prompt engineering.}
	\label{fig:2}
\end{figure*}

First, select one of the long or short prompt templates for testing. Next, choose an appropriate number of Demonstrations for assistance. Then, concatenate the claim to be verified with the selected template and Demonstrations and input this into the large language mode. Finally, the prediction result will be returned through a verbalizer. The specific process is illustrated in the Figure \ref{fig:2}.

The specific formula is as follows:

\begin{equation}\label{eq3}
p(y|x)=\prod_{j=1}^np([result]_j=V(y)|T(x)+D_m)
\end{equation}

Here, $p$ represents the entire prompt engineering method, $x$ denotes the claim to be verified, $y$ denotes the output result from the large language mode, $V$ represents the verbalizer, $V(y)$ denotes the final predicted result after verbalizer, $T$ denotes the prompt template, and $D_m$ indicates the use of $m$ samples in the in-context learning method, where $m \in \mathbb{N}$.

\section{Experiments and Results Analysis}

\subsection{Dataset Overview}

The dataset used in this study originates from the English-language portion of the research described in this paper \cite{alam2020fighting}. The data were collected from tweets containing keywords related to `COVID-19' between January 2020 and March 2021. The dataset integrates the perspectives and interests of journalists, fact-checkers, social media platforms, policymakers, and society, providing a comprehensive overview of fact-checking information in the post-pandemic era. This dataset is binary classified, with labels `Yes' and `No,' indicating whether a piece of text is worth fact-checking. The training set consists of 22,501 instances, the validation set of 1,032 instances, and the test set of 318 instances. The detailed information of the dataset is provided in Table \ref{tab:2} below:

\begin{table*}[htbp]
\caption{Specific parameters of the dataset.}
\begin{center}
\renewcommand{\arraystretch}{1.5}
\begin{tabular}{|l|c|c|c|c|c|c|}
\hline
\textbf{Dataset} & \textbf{\makecell[c]{Number of \\ samples}} & \textbf{\makecell[c]{Number of \\ positive samples}} & \textbf{\makecell[c]{Number of \\ negative samples}} & \textbf{\makecell[c]{Maximum \\ claim length}} & \textbf{\makecell[c]{Minimum \\ claim length}} & \textbf{\makecell[c]{Average \\ claim length}} \\
\hline
\multirow{2}{*}{Training set}     & \multirow{2}{*}{22501} & \multirow{2}{*}{5413} & \multirow{2}{*}{17088} & \multirow{2}{*}{833} & \multirow{2}{*}{6}  & \multirow{2}{*}{97} \\
& & & & & & \\
\hline
\multirow{2}{*}{Verification set} & \multirow{2}{*}{1032}  & \multirow{2}{*}{238}  & \multirow{2}{*}{794}   & \multirow{2}{*}{536} & \multirow{2}{*}{17} & \multirow{2}{*}{89} \\
& & & & & & \\
\hline
\multirow{2}{*}{Test set}         & \multirow{2}{*}{318}   & \multirow{2}{*}{108}  & \multirow{2}{*}{210}   & \multirow{2}{*}{324} & \multirow{2}{*}{18} & \multirow{2}{*}{67} \\
& & & & & & \\
\hline
\end{tabular}
\end{center}
\label{tab:2}
\end{table*}

\subsection{Prompt Tuning}

This study utilizes the prompt method within the parameter-efficient fine-tuning approach to enable the underlying large language model to be applied to fact-check-worthiness estimation better. This method concatenates learnable prompt parameters $P_{e}\in\mathbb{R}^{p\times e}$ as a prefix to the original input embeddings $X_{e}\in\mathbb{R}^{p\times e}$  of length $e$, resulting in $[P_e;X_e]\in\mathbb{R}^{(p+n)\times e}$, where $p$ is the length of the prompt and $e$ is the size of the embedding dimension. During training, only the added prompt layer is optimized, with the weights of the underlying model frozen, allowing for separate training and updating of the prompt parameters. Using the Qwen2.0-7B model as an example, which has 7,615,688,192 parameters, the Prompt Tuning method requires training only 71,680 parameters, accounting for 0.0009\% of the total model parameters.

\subsection{Experiment Setup}

This study selects the Qwen series of local models and the GPT series of models as the baseline models. Considering the limitations of the experimental environment and computational resources, three models with different parameter sizes—0.5B, 1.8B, and 7B—were chosen from the Qwen 1.5 series; three models with different parameter sizes—0.5B, 1.5B, and 7B—were chosen from the Qwen 2 series; and four different models—3.5, 4, 4o-mini, and 4o—were chosen from the GPT series. Fine-tuning was performed using the hyperparameters specified in Table \ref{tab:3}. All experiments were conducted on a single NVIDIA 3070 GPU, a single NVIDIA 4090 GPU, and a single NVIDIA A800 GPU. Experiments involving the GPT series models utilized the API provided by OpenAI directly.

\begin{table}[htbp]
\caption{Hyperparameters for prompt tuning.}
\begin{center}
\renewcommand{\arraystretch}{1.2}
\begin{tabular}{|c|c|c|c|}
\hline
\textbf{Learning rate} & \textbf{Number of epochs} & \textbf{Batch size} & \textbf{Optimizer} \\
\hline
\multirow{2}{*}{0.00003} & \multirow{2}{*}{5} & \multirow{2}{*}{8} & \multirow{2}{*}{AdamW} \\
& & & \\
\hline
\end{tabular}
\end{center}
\label{tab:3}
\end{table}

\subsection{Results and Analysis of Model Experiments}

Since this experiment involves a binary classification task, the selection of evaluation metrics cannot rely solely on accuracy. We use the F1 score as the primary metric to evaluate model performance. To provide a more comprehensive analysis and presentation of the data, we also present the recall and precision metrics. All results displayed below are averages from three rounds of experiments and are rounded to four decimal places.

\subsubsection{Comparison of Model Parameter Sizes and Template Lengths}

\begin{figure*}[]
	\centering
	\includegraphics[width=.9\textwidth]{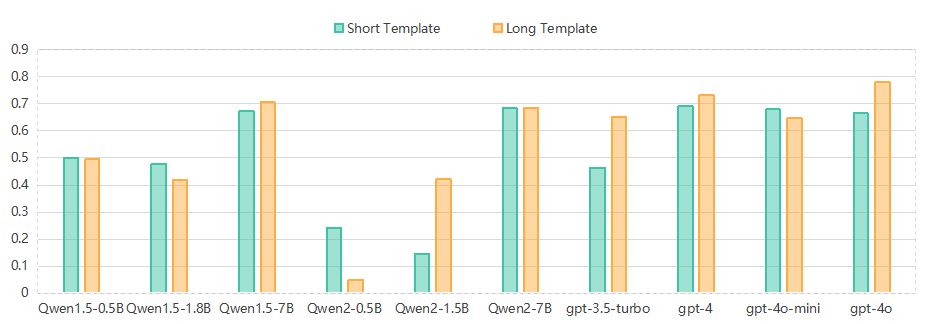}
	\caption{Comparison of model parameter sizes and template lengths.}
	\label{fig:3}
\end{figure*}

To investigate the impact of different parameter sizes on model performance, this study first tested the locally deployed Qwen series models and the GPT series models using both the long and short templates in prompt engineering. The testing methodology employed a 0-shot sampling method, meaning no Demonstrations were used, and experiments relied solely on the prompt templates. The specific results are shown in Table \ref{tab:4} and Figure \ref{fig:3}.

\begin{table*}[htbp]
\caption{Comparison of model parameter sizes and template lengths.}
\label{tab:4}
\begin{center}
\renewcommand{\arraystretch}{1.5}
\setlength{\tabcolsep}{15pt}
\begin{tabular}{|l|l|l|l|l|l|}
\hline
\textbf{Model} & \textbf{Template} & \textbf{F1 score} & \textbf{Recall} & \textbf{Precision} & \textbf{Accuracy} \\
\hline
Qwen1.5-0.5B  & Short    & 0.5013   & 0.9352 & 0.3424    & 0.3679   \\
\hline
Qwen1.5-0.5B  & Long     & 0.4950   & 0.9167 & 0.3390    & 0.3648   \\
\hline
Qwen1.5-1.8B  & Short    & 0.4774   & 0.8796 & 0.3276    & 0.3459   \\
\hline
Qwen1.5-1.8B  & Long     & 0.4198   & 0.5093 & 0.3571    & 0.5220   \\
\hline
Qwen1.5-7B    & Short    & 0.6726   & 0.6944 & 0.6522    & 0.7704   \\
\hline
Qwen1.5-7B    & Long     & 0.7059   & 0.8333 & 0.6122    & 0.7642   \\
\hline
Qwen2-0.5B    & Short    & 0.2414   & 0.1944 & 0.3182    & 0.5849   \\
\hline
Qwen2-0.5B    & Long     & 0.0484   & 0.0278 & 0.1875    & 0.6289   \\
\hline
Qwen2-1.5B    & Short    & 0.1452   & 0.0833 & 0.5625    & 0.6667   \\
\hline
Qwen2-1.5B    & Long     & 0.4211   & 0.2963 & 0.7273    & 0.7233   \\
\hline
Qwen2-7B      & Short    & 0.6836   & 0.6836 & 0.6836    & 0.6836   \\
\hline
Qwen2-7B      & Long     & 0.6868   & 0.8426 & 0.5796    & 0.7390   \\
\hline
GPT-3.5-turbo & Short    & 0.4639   & 0.3272 & 0.7982    & 0.7432   \\
\hline
GPT-3.5-turbo & Long     & 0.6508   & 0.5988 & 0.7131    & 0.7820   \\
\hline
GPT-4         & Short    & 0.6913   & 0.6389 & 0.7531    & 0.8061   \\
\hline
GPT-4         & Long     & 0.7320   & 0.6574 & 0.8262    & 0.8365   \\
\hline
GPT-4o-mini   & Short    & 0.6796   & 0.9722 & 0.5224    & 0.6887   \\
\hline
GPT-4o-mini   & Long     & 0.6477   & 0.5278 & 0.8382    & 0.8050   \\
\hline
GPT-4o        & Short    & 0.6653   & 0.5401 & 0.8664    & 0.8155   \\
\hline
GPT-4o        & Long     & 0.7803   & 0.8056 & 0.7565    & 0.8459   \\
\hline
\end{tabular}
\end{center}
\end{table*}

Based on the experimental results, we begin our analysis with the models of smaller parameter sizes. In the field of fact-check-worthiness estimation, models with smaller parameter sizes do not perform well. For instance, the F1 scores for models with 0.5B, 1.5B, and 1.8B parameters are all less than 0.500. We attribute this to the fact that these models have insufficient parameter sizes and speculate that the training data for these models do not include knowledge relevant to fact-checking. Consequently, in the simpler short templates, the models do not understand what constitutes `worth checking.' In the case of longer templates, the poor handling of complex tasks by models with smaller parameter sizes means they fail to grasp the concept of `fact-checking,' and the additional prompts actually interfere with the models’ reasoning. Except for Qwen2-1.5B, the performance of the other three models with parameter sizes less than 2B declined when transitioning from the short to the long template.

When the model parameter size reaches 7B, the performance improves significantly, with both models achieving F1 scores greater than 0.6700. Furthermore, when transitioning from the short to the long template, the performance of both models shows some degree of improvement. Although the short template performance of the Qwen1.5 model is inferior to that of the Qwen2 model, the improvement with the long template is much greater in the former, with its F1 score exceeding 0.7000, making it the best-performing local model.

For state-of-the-art GPT models, fact-check-worthiness estimation is also not a trivial problem. The worst performance was observed in the GPT-3.5 model when using short templates, with an F1 score of 0.4639. clearly indicating that GPT-3.5 did not understand the concept of 'fact-checking.' However, after switching to the long template, its F1 score increased to 0.6508, showing an improvement of 0.1869. The latest replacement model for GPT-3.5, GPT-4o-mini, achieved F1 scores of 0.6796 and 0.6477 with the short and long templates, respectively, indicating a decline in performance. The most powerful models, GPT-4 and GPT-4o, both demonstrated good performance with the short template, and their effectiveness improved when using the long template. The F1 score for GPT-4o reached 0.7803, making it the best model in this comparative experiment.

Notably, the performance of the Qwen1.5-7B model using the long template exceeded the optimal F1 scores of GPT-3.5 and GPT-4o-mini, and it also outperformed GPT-4 and GPT-4o using the short template. This preliminary finding suggests that local open-source models can hold their own in fact-check-worthiness estimation.

\subsubsection{Comparison of In-Context Learning}

This study tests the in-context learning method to assess whether the proposed method performs better when a limited number of contextual examples are available. Specifically, $n$ samples are randomly selected from the original training set to form a small-sample prompt dataset, which is then incorporated into the prompt template as part of the prompt engineering process. This experiment tested different context sizes with $n$ values set to 0, 1, 3, 5, and 10. When $n$ is 0, all models were tested using the long template prompting method. The specific results are shown in Table \ref{tab:6} in the Appendix and Figure \ref{fig:4}.

\begin{figure*}[]
	\centering
	\includegraphics[width=.9\textwidth]{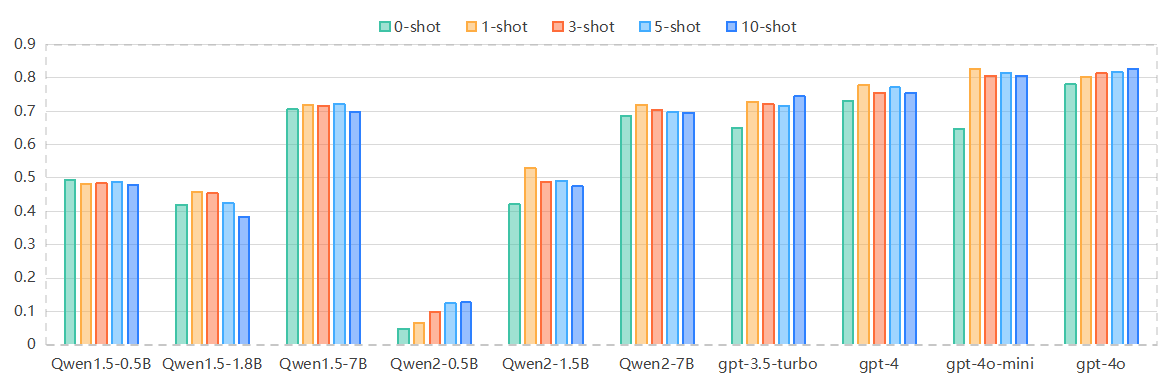}
	\caption{Comparison of in-context learning results.}
	\label{fig:4}
\end{figure*}

According to the experimental results, nine of ten models showed improved performance after using in-context learning, except Qwen1.5-0.5B. Among the Qwen series models, the most significant improvement was observed in Qwen2-1.5B, where the F1 score increased from 0.4211 to 0.5303, representing an improvement of approximately 20.59\%. In the GPT series models, the F1 score of GPT-4o-mini increased from 0.6477 to 0.8262, representing an improvement of approximately 21.60\%. As the number of prompt samples increased, not all models saw a corresponding increase in performance. For example, models such as Qwen1.5-0.5B, Qwen2-7B, and GPT4 saw their F1 scores decrease as the number of prompt samples increased. We believe that adding many prompt samples makes the overall input cumbersome, preventing the model from effectively learning from these examples. Additionally, Liu et al. \cite{liu2024lost} noted that performance is typically highest when key information appears at the beginning or end of the input context, implying that simply increasing the number of prompt samples does not necessarily enhance the model's fact-check-worthiness estimation performance. However, GPT-4o is an exception, as its classification performance is proportional to the number of prompt samples. We speculate that this may be due to optimizations in data extraction by GPT-4o, enabling it to correctly identify training samples within the prompt text and learn which texts are suitable for fact-checking.

The experimental results show that the GPT-4o model achieved the best performance in the 10-shot scenario, with an F1 score of 0.8268. Similarly, the GPT-4o-mini model in the 1-shot scenario achieved an F1 score of 0.8217. From the perspective of economic costs and response times, the GPT-4o-mini model is the most suitable choice for addressing the fact-check-worthiness estimation problem. In the 5-shot scenario, the Qwen1.5-7B model emerged as the best local model, achieving an F1 score of 0.7217, which is comparable to the performance of the GPT-3.5 model. This indicates that simple few-shot learning methods have limited effectiveness in prompting models with smaller parameter sizes, preventing them from reaching the performance levels of the GPT series models.

\subsubsection{Comparison of Prompt Tuning}

Prompt tuning was performed on the Qwen1.5 series models with parameter sizes of 0.5B, 1.8B, and 7B, as well as the Qwen2 series models with parameter sizes of 0.5B, 1.5B, and 7B. For ease of comparison, the 'best' row and 'best' legend in Figure \ref{fig:5} and Table \ref{tab:7} in the appendix represent the best performance achieved by the models without any prompt tuning.

\begin{figure*}[]
	\centering
	\includegraphics[width=.9\textwidth]{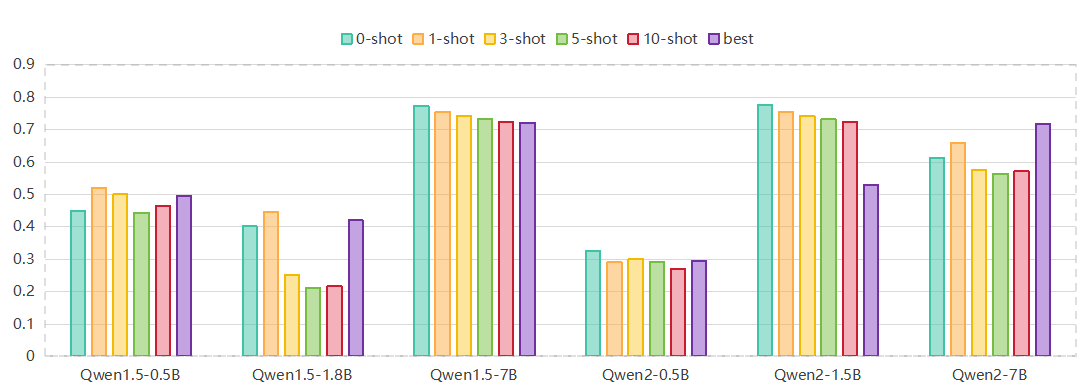}
	\caption{Comparison of prompt tuning.}
	\label{fig:5}
\end{figure*}

From the experimental results, it can be seen that prompt tuning indeed enhances the performance of some models. Except for Qwen2-7B, the performance of the other five models improved. However, even after fine-tuning, the models still exhibit poor adaptability to in-context learning, with performance declining as the number of samples increases. The Qwen2-1.5B model showed a significant improvement after fine-tuning, with its score in the 0-shot scenario increasing from 0.5303 to 0.7774, even surpassing the F1 score of 0.7737 achieved by Qwen1.5-7B. It achieved the best performance with a relatively smaller parameter size.

\subsubsection{Comparison with Baseline}

To validate the effectiveness of the proposed method, we compare the best-performing fine-tuned model with classic classification models and the current state-of-the-art GPT models. The following methods were selected as baseline approaches:

\begin{itemize}

\item Majority: This baseline uses a Dummy classifier, which does not provide any insights into the data and classifies given data using simple rules. Typically, this classifier serves as a simple baseline for comparison with other classifiers.

\item Random: Random classification, where each result is predicted randomly with a 50\% probability.

\item N-grams: Features are constructed by calculating the frequency or probability of continuous word sequences (N-grams) in the text, and machine learning algorithms are then applied for document classification.

\item Bert: A classic pre-trained text classification model trained through masked language modeling, capable of learning rich bidirectional textual context representations, achieving state-of-the-art performance in numerous natural language processing tasks.

\item GPT-3.5: The GPT-3.5-turbo large language model developed by OpenAI, leading the wave of large language models. Here, we consider the GPT-3.5-turbo model in a 10-shot scenario.

\item GPT-4: The GPT-4-turbo-preview large language model developed by OpenAI is the most comprehensive large language model currently available. Here, we consider the GPT-4 model in a 1-shot scenario.

\item GPT-4o-mini: The lightweight large language model GPT-4o-mini introduced by OpenAI, characterized by low computational resource requirements and high generation quality, is suitable for applications that balance efficiency and performance. Here, we consider the GPT-4o-mini model in a 1-shot scenario.

\item GPT-4o: The GPT-4o large language model developed by OpenAI is known for its excellent understanding and generation capabilities, suitable for high-precision task handling, and considered one of the advanced models achieving the best balance between generation quality and computational efficiency. Here, we consider the GPT-4o model in a 10-shot scenario.

\item Ours: The best-performing local model selected, which is the Qwen2-1.5B model after prompt tuning in a 0-shot scenario.

\end{itemize}

The results of the comparative experiments between the proposed method and the eight baseline methods are shown in Table \ref{tab:5} and Figure \ref{fig:6} below.

\begin{table}[htbp]
\caption{Comparison of baseline.}
\label{tab:5}
\begin{center}
\renewcommand{\arraystretch}{1.2}
\begin{tabular}{|l|l|l|l|l|}
\hline
\textbf{Model} & \textbf{F1 score} & \textbf{Recall} & \textbf{Precision} & \textbf{Accuracy} \\
\hline
Majority      & 0.0000   & 0.0000 & 0.0000    & 0.6604   \\
\hline
Random        & 0.3883   & 0.4907 & 0.3212    & 0.4748   \\
\hline
N-grams        & 0.5988   & 0.4630 & 0.8475    & 0.7893   \\
\hline
Bert          & 0.7458   & 0.6111 & 0.9565    & 0.8585   \\
\hline
gpt-3.5-turbo & 0.7450   & 0.8025 & 0.6954    & 0.8134   \\
\hline
gpt-4         & 0.7788   & 0.8148 & 0.7458    & 0.8428   \\
\hline
gpt-4o-mini   & 0.8262   & 0.8580 & 0.7968    & 0.8774   \\
\hline
gpt-4o        & 0.8268   & 0.8611 & 0.7951    & 0.8774   \\
\hline
Ours          & 0.7774   & 0.8655 & 0.7055    & 0.8857   \\
\hline
\end{tabular}
\end{center}
\end{table}

\begin{figure*}[]
	\centering
	\includegraphics[width=.9\textwidth]{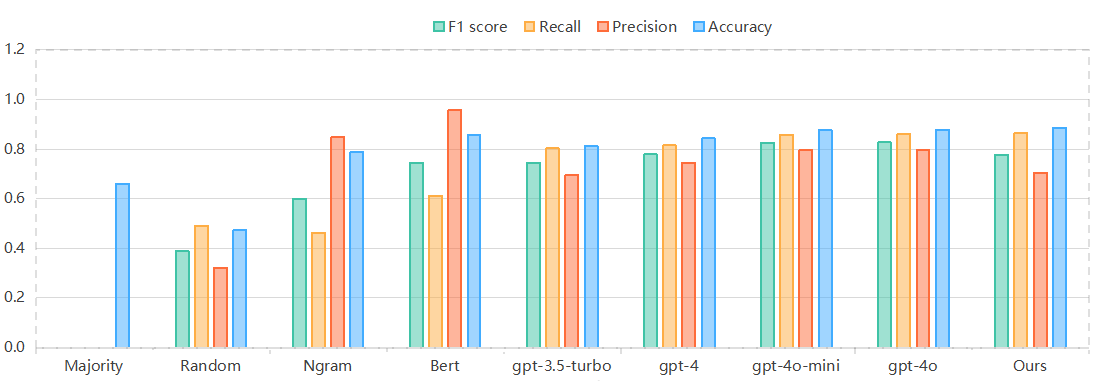}
	\caption{Comparison of baseline.}
	\label{fig:6}
\end{figure*}

Experimental validation has shown that the proposed method demonstrates promising results in the task of fact-check-worthiness estimation, achieving an F1 score of 0.7774. This score surpasses all classical methods in the baselines and outperforms the GPT-3.5 model while being comparable to the GPT-4 model. However, there remains a gap when compared to the latest GPT-4o-mini and GPT-4o models. Nonetheless, our method ranks first among all models in terms of accuracy. Given its parameter size of only 1.5B, the proposed method still holds certain advantages when considering the overall performance.

\section{Conclusion and Future Work}

This paper proposes a fact-check-worthiness estimation method based on prompt tuning. The method combines the design of prompt engineering templates, the use of in-context learning, and prompt tuning to improve the accuracy of model classification. Extensive experiments on public datasets demonstrate that the proposed method effectively enhances the accuracy of fact-check-worthiness estimation. Despite its smaller parameter size and lower hardware requirements, the performance achieved is comparable to that of the GPT-4 model.

Furthermore, the method proposed in this paper has room for improvement. In terms of prompt template design, we have only created two types of templates, and there may be better designs that we have not yet considered. Additionally, the chain-of-thought template design approach could also be applicable to this problem. Regarding the fine-tuning method, we chose only prompt tuning, and methods such as LoRA, Prefix Tuning, and P-Tuning have not been validated. These will be the focus of our next steps. In the selection of local models, we only evaluated the Qwen series models, whereas others like LLaMA and ChatGLM have not been tested. Due to computational resource constraints, we only selected models with a maximum parameter size of 7B, whereas larger models with 14B or 72B parameters might exhibit better performance. In terms of dataset selection, we only used English datasets, and tasks involving fact-check-worthiness in other languages, such as Chinese and German, have not been assessed.

Finally, we believe that future work will uncover deeper potential in large language models for fact-checking tasks.

\bibliographystyle{IEEEtran}
\bibliography{main}
\vspace{12pt}

\appendices
\section{Supplementary data}
\label{appendixA}
To avoid disrupting the reader's experience, we have placed the two lengthy tables, Table \ref{tab:6} and Table \ref{tab:7}, in this appendix.

\begin{table*}[htbp]
\caption{Comparison of in-context learning results}
\label{tab:6}
\begin{center}
\renewcommand{\arraystretch}{1.3}
\setlength{\tabcolsep}{15pt}
\begin{tabular}{|c|c|c|c|c|c|}
\hline
\textbf{Model} & \textbf{Demonstration} & \textbf{F1 score} & \textbf{Recall} & \textbf{Precision} & \textbf{Accuracy} \\
\hline
Qwen1.5-0.5B  & 0-shot   & 0.4950   & 0.9167 & 0.3390    & 0.3648   \\
\hline
Qwen1.5-0.5B  & 1-shot   & 0.4824   & 0.8457 & 0.3375    & 0.3836   \\
\hline
Qwen1.5-0.5B  & 3-shot   & 0.4837   & 0.8889 & 0.3322    & 0.3553   \\
\hline
Qwen1.5-0.5B  & 5-shot   & 0.4877   & 0.8889 & 0.3360    & 0.3658   \\
\hline
Qwen1.5-0.5B  & 10-shot  & 0.4797   & 0.8210 & 0.3389    & 0.3952   \\
\hline
Qwen1.5-1.8B  & 0-shot   & 0.4198   & 0.5093 & 0.3571    & 0.5220   \\
\hline
Qwen1.5-1.8B  & 1-shot   & 0.4569   & 0.5401 & 0.3969    & 0.5671   \\
\hline
Qwen1.5-1.8B  & 3-shot   & 0.4538   & 0.5000 & 0.4159    & 0.5912   \\
\hline
Qwen1.5-1.8B  & 5-shot   & 0.4252   & 0.4475 & 0.4049    & 0.5891   \\
\hline
Qwen1.5-1.8B  & 10-shot  & 0.3841   & 0.3580 & 0.4143    & 0.6101   \\
\hline
Qwen1.5-7B    & 0-shot   & 0.7059   & 0.8333 & 0.6122    & 0.7642   \\
\hline
Qwen1.5-7B    & 1-shot   & 0.7198   & 0.8765 & 0.6110    & 0.7683   \\
\hline
Qwen1.5-7B    & 3-shot   & 0.7171   & 0.7747 & 0.6676    & 0.7925   \\
\hline
Qwen1.5-7B    & 5-shot   & 0.7217   & 0.7284 & 0.7152    & 0.8092   \\
\hline
Qwen1.5-7B    & 10-shot  & 0.6988   & 0.6698 & 0.7309    & 0.8040   \\
\hline
Qwen2-0.5B    & 0-shot   & 0.0484   & 0.0278 & 0.1875    & 0.6289   \\
\hline
Qwen2-0.5B    & 1-shot   & 0.0661   & 0.0370 & 0.3077    & 0.6447   \\
\hline
Qwen2-0.5B    & 3-shot   & 0.0986   & 0.0648 & 0.2059    & 0.5975   \\
\hline
Qwen2-0.5B    & 5-shot   & 0.1241   & 0.0833 & 0.2432    & 0.6006   \\
\hline
Qwen2-0.5B    & 10-shot  & 0.1268   & 0.0833 & 0.2647    & 0.6101   \\
\hline
Qwen2-1.5B    & 0-shot   & 0.4211   & 0.2963 & 0.7273    & 0.7233   \\
\hline
Qwen2-1.5B    & 1-shot   & 0.5303   & 0.6481 & 0.4487    & 0.6101   \\
\hline
Qwen2-1.5B    & 3-shot   & 0.4871   & 0.4871 & 0.4871    & 0.4871   \\
\hline
Qwen2-1.5B    & 5-shot   & 0.4919   & 0.5648 & 0.4363    & 0.6038   \\
\hline
Qwen2-1.5B    & 10-shot  & 0.4750   & 0.5062 & 0.4480    & 0.6205   \\
\hline
Qwen2-7B      & 0-shot   & 0.6868   & 0.8426 & 0.5796    & 0.7390   \\
\hline
Qwen2-7B      & 1-shot   & 0.7186   & 0.9414 & 0.5811    & 0.7495   \\
\hline
Qwen2-7B      & 3-shot   & 0.7038   & 0.8765 & 0.5879    & 0.7495   \\
\hline
Qwen2-7B      & 5-shot   & 0.6973   & 0.8673 & 0.5830    & 0.7442   \\
\hline
Qwen2-7B      & 10-shot  & 0.6960   & 0.8272 & 0.6010    & 0.7547   \\
\hline
gpt-3.5-turbo & 0-shot   & 0.6508   & 0.5988 & 0.7131    & 0.7820   \\
\hline
gpt-3.5-turbo & 1-shot   & 0.7279   & 0.7099 & 0.7470    & 0.8197   \\
\hline
gpt-3.5-turbo & 3-shot   & 0.7224   & 0.7438 & 0.7033    & 0.8061   \\
\hline
gpt-3.5-turbo & 5-shot   & 0.7160   & 0.7191 & 0.7131    & 0.8061   \\
\hline
gpt-3.5-turbo & 10-shot  & 0.7450   & 0.8025 & 0.6954    & 0.8134   \\
\hline
gpt-4         & 0-shot   & 0.7320   & 0.6574 & 0.8262    & 0.8365   \\
\hline
gpt-4         & 1-shot   & 0.7788   & 0.8148 & 0.7458    & 0.8428   \\
\hline
gpt-4         & 3-shot   & 0.7556   & 0.7870 & 0.7265    & 0.8270   \\
\hline
gpt-4         & 5-shot   & 0.7713   & 0.7963 & 0.7478    & 0.8396   \\
\hline
gpt-4         & 10-shot  & 0.7550   & 0.7654 & 0.7453    & 0.8312   \\
\hline
gpt-4o        & 0-shot   & 0.7803   & 0.8056 & 0.7565    & 0.8459   \\
\hline
gpt-4o        & 1-shot   & 0.8029   & 0.8611 & 0.7522    & 0.8564   \\
\hline
gpt-4o        & 3-shot   & 0.8139   & 0.8642 & 0.7693    & 0.8658   \\
\hline
gpt-4o        & 5-shot   & 0.8180   & 0.8673 & 0.7741    & 0.8690   \\
\hline
gpt-4o        & 10-shot  & 0.8268   & 0.8611 & 0.7951    & 0.8774   \\
\hline
gpt-4o-mini   & 0-shot   & 0.6477   & 0.5278 & 0.8382    & 0.8050   \\
\hline
gpt-4o-mini   & 1-shot   & 0.8262   & 0.8580 & 0.7968    & 0.8774   \\
\hline
gpt-4o-mini   & 3-shot   & 0.8068   & 0.8765 & 0.7477    & 0.8574   \\
\hline
gpt-4o-mini   & 5-shot   & 0.8148   & 0.9167 & 0.7335    & 0.8585   \\
\hline
gpt-4o-mini   & 10-shot  & 0.8071   & 0.9228 & 0.7173    & 0.8501   \\
\hline
\end{tabular}
\end{center}
\end{table*}

\begin{table*}[htbp]
\caption{Comparison of prompt tuning.}
\label{tab:7}
\begin{center}
\renewcommand{\arraystretch}{1.3}
\setlength{\tabcolsep}{15pt}
\begin{tabular}{|c|c|c|c|c|c|}
\hline
\textbf{Model} & \textbf{Demonstration} & \textbf{F1 score} & \textbf{Recall} & \textbf{Precision} & \textbf{Accuracy} \\
\hline
Qwen1.5-0.5B & 0-shot        & 0.4497   & 0.3395 & 0.6664    & 0.7180   \\
\hline
Qwen1.5-0.5B & 1-shot        & 0.5194   & 0.6821 & 0.4196    & 0.5723   \\
\hline
Qwen1.5-0.5B & 3-shot        & 0.5003   & 0.7130 & 0.3855    & 0.5168   \\
\hline
Qwen1.5-0.5B & 5-shot        & 0.4434   & 0.6265 & 0.3432    & 0.4654   \\
\hline
Qwen1.5-0.5B & 10-shot       & 0.4654   & 0.6759 & 0.3549    & 0.4727   \\
\hline
Qwen1.5-0.5B & best          & 0.4950   & 0.9167 & 0.3390    & 0.3648   \\
\hline
Qwen1.5-1.8B & 0-shot        & 0.4017   & 0.4352 & 0.3730    & 0.5597   \\
\hline
Qwen1.5-1.8B & 1-shot        & 0.4463   & 0.6235 & 0.3475    & 0.4738   \\
\hline
Qwen1.5-1.8B & 3-shot        & 0.2517   & 0.2037 & 0.3302    & 0.5870   \\
\hline
Qwen1.5-1.8B & 5-shot        & 0.2116   & 0.1667 & 0.2911    & 0.5797   \\
\hline
Qwen1.5-1.8B & 10-shot       & 0.2169   & 0.1605 & 0.3361    & 0.6090   \\
\hline
Qwen1.5-1.8B & best          & 0.4198   & 0.5093 & 0.3571    & 0.5220   \\
\hline
Qwen1.5-7B   & 0-shot        & 0.7737   & 0.8704 & 0.6963    & 0.8270   \\
\hline
Qwen1.5-7B   & 1-shot        & 0.7547   & 0.8920 & 0.6543    & 0.8029   \\
\hline
Qwen1.5-7B   & 3-shot        & 0.7429   & 0.8025 & 0.6920    & 0.8113   \\
\hline
Qwen1.5-7B   & 5-shot        & 0.7322   & 0.7346 & 0.7302    & 0.8176   \\
\hline
Qwen1.5-7B   & 10-shot       & 0.7246   & 0.6944 & 0.7575    & 0.8208   \\
\hline
Qwen1.5-7B   & best          & 0.7217   & 0.7284 & 0.7152    & 0.8092   \\
\hline
Qwen2-0.5B   & 0-shot        & 0.3270   & 0.6106 & 0.2233    & 0.4202   \\
\hline
Qwen2-0.5B   & 1-shot        & 0.2904   & 0.3964 & 0.2292    & 0.5536   \\
\hline
Qwen2-0.5B   & 3-shot        & 0.3006   & 0.4216 & 0.2337    & 0.5475   \\
\hline
Qwen2-0.5B   & 5-shot        & 0.2915   & 0.4160 & 0.2244    & 0.5336   \\
\hline
Qwen2-0.5B   & 10-shot       & 0.2714   & 0.3754 & 0.2127    & 0.5355   \\
\hline
Qwen2-0.5B   & best          & 0.2941   & 0.2778 & 0.3125    & 0.5472   \\
\hline
Qwen2-1.5B   & 0-shot        & 0.7774   & 0.8655 & 0.7055    & 0.8857   \\
\hline
Qwen2-1.5B   & 1-shot        & 0.7547   & 0.8920 & 0.6543    & 0.8029   \\
\hline
Qwen2-1.5B   & 3-shot        & 0.7429   & 0.8025 & 0.6920    & 0.8113   \\
\hline
Qwen2-1.5B   & 5-shot        & 0.7322   & 0.7346 & 0.7302    & 0.8176   \\
\hline
Qwen2-1.5B   & 10-shot       & 0.7246   & 0.6944 & 0.7575    & 0.8208   \\
\hline
Qwen2-1.5B   & best          & 0.5303   & 0.6481 & 0.4487    & 0.6101   \\
\hline
Qwen2-7B     & 0-shot        & 0.6126   & 0.9202 & 0.4591    & 0.7316   \\
\hline
Qwen2-7B     & 1-shot        & 0.6581   & 0.9678 & 0.9678    & 0.9678   \\
\hline
Qwen2-7B     & 3-shot        & 0.5763   & 0.9706 & 0.4098    & 0.6709   \\
\hline
Qwen2-7B     & 5-shot        & 0.5647   & 0.9874 & 0.3954    & 0.6489   \\
\hline
Qwen2-7B     & 10-shot       & 0.5717   & 0.9804 & 0.4035    & 0.6612   \\
\hline
Qwen2-7B     & best          & 0.7186   & 0.9414 & 0.5811    & 0.7495   \\
\hline
\end{tabular}
\end{center}
\end{table*}

\end{document}